\pdfoutput=1

\documentclass[11pt]{article}

\usepackage{acl}
\usepackage{times}
\usepackage{latexsym}
\usepackage{amsmath}
\usepackage{kotex}
\usepackage{devanagari}
\usepackage{lipsum}
\usepackage{graphics}
\usepackage{placeins}
\usepackage{graphicx}

\usepackage[T1]{fontenc}

\usepackage[utf8]{inputenc}

\usepackage{microtype}

\title{Evaluating Gender Bias in Hindi-English Machine Translation}

\author{Gauri Gupta* \\
  Manipal Institute of Technology \\
  MAHE, Manipal, 576104 \\
  \texttt{gaurigupta.315@gmail.com} \\\And
  Krithika Ramesh*\\
  Manipal Institute of Technology \\
  MAHE, Manipal, 576104 \\
  \texttt{kramesh.tlw@gmail.com} \\ \AND
  Sanjay Singh\\
  Manipal Institute of Technology \\
  MAHE, Manipal, 576104 \\
  \texttt{sanjay.singh@manipal.edu}}

\begin{document}
\maketitle
\begin{abstract}
With language models being deployed increasingly in the real world, it is essential to address the issue of the fairness of their outputs. The word embedding representations of these language models often implicitly draw unwanted associations that form a social bias within the model. The nature of gendered languages like Hindi, poses an additional problem to the quantification and mitigation of bias, owing to the change in the form of the words in the sentence, based on the gender of the subject. Additionally, there is sparse work done in the realm of measuring and debiasing systems for Indic languages. In our work, we attempt to evaluate and quantify the gender bias within a Hindi-English machine translation system. We implement a modified version of the existing TGBI metric based on the grammatical considerations for Hindi. We also compare and contrast the resulting bias measurements across multiple metrics for pre-trained embeddings and the ones learned by our machine translation model.
\end{abstract}

\section{Introduction}
There has been a recent increase in the studies on gender bias in natural language processing considering bias in word embeddings, bias amplification, and methods to evaluate bias \cite{savoldi2021gender}, with some evaluation methods introduced primarily to measure gender bias in MT systems. In MT systems, bias can be identified as the cause of the translation of gender-neutral sentences into gendered ones. There has been little work done for bias in language models for Hindi, and to the best of our knowledge, there has been no previous work that measures and analyses bias for MT of Hindi. Our approach uses two existing and broad frameworks for assessing bias in MT, including the Word Embedding Fairness Evaluation \cite{ijcai2020-60} and the Translation Gender Bias Index \cite{cho-etal-2019-measuring} on Hindi-English MT systems. We modify some of the existing procedures within these metrics required for compatibility with Hindi grammar.
This paper contains the following contributions:
\begin{enumerate}
\item Construction of an equity evaluation corpus (EEC) \cite{DBLP:journals/corr/abs-1805-04508} for Hindi of size 26370 utterances using 1558 sentiment words and 1100 occupations following the guidelines laid out in \citet{cho-etal-2019-measuring}.
\item Evaluation of gender bias in MT systems for Indic languages. 
\item An emphasis on a shift towards inclusive models and metrics. The paper is also demonstrative of language that should be used in NLP papers working on gender bias. 
\end{enumerate}
All our codes and files are publicly available.\footnote{https://github.com/stolenpyjak/hi-en-bias-eval}

\section{Related Work}
The prevalence of social bias within a language model is caused by it inadvertently drawing unwanted associations within the data. Previous works that have addressed tackling bias include \citet{DBLP:journals/corr/BolukbasiCZSK16a}, which involved the use of multiple gender-definition pairs and principal component analysis to infer the direction of the bias. In order to mitigate the bias, each word vector had its projection on this subspace subtracted from it. However, this does not entirely debias the word vectors, as noted in \citet{gonen-goldberg-2019-lipstick}.

There have been various attempts to measure the bias in existing language models. \citet{huang-etal-2020-reducing} measure bias based on whether the sentiment of the generated text would alter if there were a change in entities such as the occupation, gender, etc. \citet{kurita-etal-2019-measuring} performed experiments on evaluating the bias in BERT using the Word Embedding Association Test (WEAT) as a baseline for their own metric, which involved calculating the mean of the log probability bias score for each attribute.

Concerning the measurement of bias in existing MT systems, \citet{stanovsky-etal-2019-evaluating} came up with a method to evaluate gender bias for 8 target languages automatically. Their experiments aligned translated text with the source text and then mapped the English entity (source) to the corresponding target translation, from which the gender is extracted.

Most of the focus in mitigating bias has been in English, which is not a gendered language. Languages like Hindi and Spanish contain grammatical gender, where the gender of the verbs, articles, adjectives must remain consistent with that of the gender of the noun. In \citet{zhou2019examining} a modified version of WEAT was used to measure the bias in Spanish and French, based on whether the noun was inanimate or animate, with the latter containing words like `doctor,' which have two variants for `male' and `female' each. \citet{gonen2019does} worked on addressing the problem with such inanimate nouns as well and attempted to neutralize the grammatical gender signal of these words during training by lemmatizing the context words and changing the gender of these words.

While there has been much work on quantifying and mitigating bias in many languages in NLP, the same cannot be said for Hindi and other Indic languages, possibly because they are low-resource. \citet{10.1145/3377713.3377792} was the first work in this area; they use geometric debiasing, where a bias subspace is first defined and the word is decomposed into two components, of which the gendered component is reduced. Finally, SVMs were used to classify the words and quantify the bias. 

\section{Methodology}
\subsection{Dataset and Data Preprocessing}

The trained model that we borrowed from \citet{10.1007/978-981-33-4909-4_25} was trained on the IIT-Bombay Hindi-English parallel data corpus \cite{kunchukuttan-etal-2018-iit}, which contains approximately 1.5 million examples across multiple topics. \citet{10.1007/978-981-33-4909-4_25} used back-translation to increase the performance of the existing model by training the English-Hindi model on the IIT-Bombay corpus and then subsequently used it to translate 3 million records in the WMT-14 English monolingual dataset to augment the existing parallel corpus training data. The model was trained on this back-translated data, which was split into 4 batches. 

The dataset cleaning involved removing special characters, punctuation, and other noise, and the text was subsequently converted to lowercase. Any duplicate records within the corpus were also removed, word-level tokenization was implemented, and the most frequent 50,000 tokens were retained. In the subword level tokenization, where byte-pair encoding was implemented, 50,000 subword tokens were created and added to this vocabulary.

\subsection{NMT Model Architecture}

For our experiments in building the neural machine translation model, we made use of the OpenNMT-tf \cite{klein-etal-2020-opennmt} library, with the model's configuration being borrowed from \citet{10.1007/978-981-33-4909-4_25}. The OpenNMT model made use of the Transformer architecture \cite{vaswani2017attention}, consisting of 6 layers each in the encoder and decoder architecture, with 512 hidden units in every hidden layer. The dimension of the embedding layer was set to 512, with 8 attention heads, with the LazyAdam optimizer being used to optimize model parameters. The batch size was 64 samples, and the effective batch size for each step was 384.

\begin{table*}[t]
\centering
\begin{tabular}{|c|l|l|l|l|}
\hline
\textbf{Embedding}           & \multicolumn{1}{c|}{\textbf{WEAT}} & \multicolumn{1}{c|}{\textbf{RNSB}} & \multicolumn{1}{c|}{\textbf{RND}} & \multicolumn{1}{c|}{\textbf{ECT}} \\ \hline
\textbf{NMT-English-(512D)}  & 0.326529                           & 0.018593                           & 0.065842                          & 0.540832                          \\ \hline
\textbf{w2v-google-news-300} & 0.638202                           & 0.01683                            & 0.107376                          & 0.743634                          \\ \hline
\textbf{hi-300}              & 0.273154                           & 0.02065                            & 0.168989                          & 0.844888                          \\ \hline
\textbf{NMT-Hindi-(512D)}    & 0.182402                           & 0.033457                           & 0.031325                          & 0.299023                          \\ \hline
\end{tabular}
\caption{\label{WEFE Results}
This table depicts the results for the various metrics that were used on the embeddings, and the final values based on their ranking by the Word Embedding Fairness Evaluation Framework.
}
\end{table*}

\subsection{WEFE}

The Word Embedding Fairness Evaluation framework is used to rank word embeddings using a set of fairness criteria. WEFE takes in a query, which is a pair of two sets of target words and sets of attribute words each, which are generally assumed to be characteristics related to the former.
\begin{equation}
    Q=(\{T_{women},T_{men}\},\{A_{career},A_{family}\})
\end{equation}

The WEFE ranking process takes in an input of a set of multiple queries which serve as tests across which bias is measured $Q$, a set of pre-trained word embeddings $M$, and a set of fairness metrics $F$.

\subsubsection{The Score Matrix} Assume a fairness metric $K$ is chosen from the set $F$, with a query template $s=(t,a)$, where all subqueries must satisfy this template.  Then,
\begin{equation}
    Q_{K}=Q_{1}(s)\cup Q_{2}(s) \cup... \cup Q_{r}(s)
\end{equation}

In that case, the $Q_{i}(s)$ forms the set of all subqueries that satisfy the query template. Thus, the value of $F=(m,Q)$ is computed for every pre-trained embedding $m$ that belongs to the set $M$, for each query present in the set. The matrix produced after doing this for each embedding is of the dimensions $M \times Q_{K}$.

The rankings are created by aggregating the scores for each row in the aforementioned matrix, which corresponds to each embedding. The aggregation function chosen must be consistent with the fairness metric, where the following property must be satisfied for $\leq_{F}$, where $x,x^{'},y,y^{'}$ are random values in |\&${\rm I\!R}$, then $agg(x,x^{'})\leq agg(y,y^{'})$ must hold true to be able to use the aggregation function. The result after performing this operation for every row is a vector of dimensions $1 \times M$, and we use $\leq{F}$ to create a ranking for every embedding, with a smaller score being ranked higher than lower ones. 

After performing this process for every fairness metric over each embedding $m \in M$, the resultant matrix with dimensions $M \times F$ consisting of the ranking indices of every embedding for every metric, and this allows us to compare and analyze the correlations of the different metrics for every word embedding.

\subsection{Metrics}
\subsubsection{WEAT}
The WEAT (Word Embedding Association Test) \cite{Caliskan_2017} metric, inspired by the IAT (Implicit Association Test), takes in a set of queries as its input, with the queries consisting of sets of target words, and attribute words. In our case, we have defined two sets of target words catering to the masculine and feminine gendered words, respectively. In addition to this, we have defined multiple pairs of sets of attribute words, as mentioned in the Appendix. WEAT calculates the association of the target set $T_{1}$ with the attribute set $A_{1}$ over the attribute set $A_{2}$, relative to $T_{2}$. For example, as observed in Table \ref{WEFE Results}, the masculine words tend to have a greater association with career than family than the feminine words. Thus, given a word $w$ in the word embedding:
\begin{equation}
\resizebox{0.48\textwidth}{!}{
    $d(w,A_{1},A_{2})= (mean_{x \in A_{1}} cos(w,x))-(mean_{x \in A_{2}} cos(w,x))$}
\end{equation}

The difference of the mean of the cosine similarities of a given word's embedding vector with the word embedding vectors of the attribute sets are utilized in the following equation to give an estimate of the association.
\begin{equation}
\resizebox{0.48\textwidth}{!}{$F_{WEAT}(M,Q)=\Sigma_{w \in T_{1}}\ d(w,A_{1},A_{2})-\Sigma_{w \in T_{2}}\ d(w,A_{1},A_{2})$}
\end{equation}

\subsubsection{RND}
The objective of the Relative Norm Distance (RND) \cite{Garg_2018} is to average the embedding vectors within the target set $T$, and for every attribute $a \in A$, the norm of the difference between the average target and the attribute word is calculated, and subsequently subtracted.
\begin{equation}
    \sum_{x \in A}(\left \|avg(T_{1})-x \right \|_{2}-\left \|avg(T_{2})-x \right \|_{2})
\end{equation}
The higher the value of the relative distance from the norm, the more associated the attributes are with the second target group, and vice versa.

\subsubsection{RNSB}
The Relative Negative Sentiment Bias (RNSB) \cite{sweeney-najafian-2019-transparent} takes in multiple target sets and two attribute sets and creates a query. Initially, a binary classifier is constructed, using the first attribute set $A_{1}$ as training examples for the first class, and $A_{2}$ for the second class. The classifier subsequently assigns every word $w$ a probability, which implies its association with an attribute set, i.e
\begin{equation}
    p(A_{1})=C_{(A_{1},A_{2})} (w)
\end{equation}
Here, $C_{(A_{1},A_{2})} (x)$ represents the binary classifier for any word x. The probability of the word's association with the attribute set $A_{2}$ would therefore be calculated as $1-C_{(A_{1},A_{2})} (w)$. A probability distribution $P$ is formed for every word in each of the target sets by computing this degree of association. Ideally, a uniform probability distribution $U$ should be formed, which would indicate that there is no bias in the word embeddings with respect to the two attributes selected. The less uniform the distribution is, the more the bias. We calculate the RNSB by defining the Kulback-Leibler divergence of $P$ from $U$ to assess the similarity of these distributions. 

\begin{table*}[t]
\centering
\begin{tabular}{|c|l|l|l|}
\hline
\textbf{Sentence}                       & \multicolumn{1}{c|}{\textbf{Size}} & \multicolumn{1}{c|}{\textbf{OpenNMT-tf}} & \multicolumn{1}{c|}{\textbf{Google Translate}} \\ \hline
Informal                                & 2628                               & 0.7543 (0.0315, 0.7473)                  & 0.3553 (0.2763, 0.2146)                        \\ \hline
Formal                                  & 5286                               & 0.5410 (0.0773, 0.5090)                  & 0.5464 (0.1015, 0.5066)                        \\ \hline
Impolite                                & 2628                               & 0.2127 (0.1552, 0.0966)                  & 0.2716 (0.1990, 0.1400)                        \\ \hline
Polite                                  & 2658                               & 0.9168 (0.0003, 0.9168)                  & 0.8690 (0.0052 0.8683)                         \\ \hline
\multicolumn{1}{|c|}{Positive}          & 2460                               & 0.6765 (0.0825, 0.6548)                  & 0.5819 (0.1589, 0.5329)                      \\ \hline
\multicolumn{1}{|c|}{Negative}          & 2212                               & 0.6773 (0.0641, 0.6773)                  & 0.5384 (0.15822, 0.5384)                       \\ \hline
\multicolumn{1}{|c|}{Occupation}        & 3242                               & 0.5100 (0.0453, 0.4888)                  & 0.3599 (0.1610, 0.2680)                        \\ \hline
\multicolumn{1}{|l|}{\textbf{Average:}} &                                    & \textbf{0.6127}                          & \textbf{0.5032}                                \\ \hline
\end{tabular}
\caption{\label{TGBI Results}
The values present under each MT system shows it's corresponding $P_{i} (p_{she},p_{they})$
value for each sentence set and the average TGBI value is calculated in the last row. 
}
\end{table*}

\subsubsection{ECT}
The Embedding Coherence Test \cite{DBLP:journals/corr/abs-1901-07656} compares the vectors of the two target sets $T_{1}$ and $T_{2}$, averaged over all their terms, with vectors from an attribute set $A$. It does so by computing mean vectors for each of these target sets such that:
\begin{equation}
    \mu_{i}=\frac1 {\lvert{T_{i}}\rvert } \Sigma_{t_{i} \in T_{i}}\ t_{i}
\end{equation}
After calculating the mean vectors for each target set, we compute its cosine similarity with every attribute vector $a \in A$, resulting in $s_{1}$ and $s_{2}$, which are vector representations of the similarity score for the target sets. The ECT score is computed by calculating the Spearman's rank correlation between the rank orders of $s_{1}$ and $s_{2}$, with a higher correlation implying lower bias.

\subsection{TGBI}

The Translation Gender Bias Index (TGBI) is a measure to detect and evaluate the gender bias in MT systems, introduced by \citet{cho-etal-2019-measuring}. They use Korean-English (KN-EN) translation. In \citet{cho-etal-2019-measuring}, the authors create a test set of words or phrases that are gender neutral in the source language, Korean. These lists were then translated using three different models and evaluated for bias using their evaluation scheme. The evaluation methodology proposed in the paper quantifies associations of `he,' `she,' and related gendered words present translated text. We carry out this methodology for Hindi, a gendered low-resource language in natural language processing tasks.

\subsubsection{Occupation and Sentiment Lists}
Considering all of the requirements laid out by \citet{cho-etal-2019-measuring}, we created a list of unique occupations and positive and negative sentiment in our source language, Hindi. The occupation list was generated by translating the list in the original paper. The translated lists were manually checked for errors and for the removal of any spelling, grammatical errors, and gender associations within these lists by native Hindi speakers. The sentiment lists were generated using the translation of existing English sentiment lists \cite{10.1145/1060745.1060797, 10.1145/1014052.1014073} and then manually checked for errors by the authors. This method of generation of sentiment lists in Hindi using translation was also seen in \citet{bakliwal-etal-2012-hindi}. 

The total lists of unique occupations and positive and negative sentiment words come out to be 1100, 820 and 738 in size respectively. These lists have also been made available online.\footnote{https://github.com/stolenpyjak/hi-en-bias-eval}

\subsubsection{Pronouns and Suffixes}
Hindi, unlike Korean, does not have gender-specific pronouns in the third person. \citet{cho-etal-2019-measuring} considered 그 사람 (ku salam), `the person' as a formal gender-neutral pronoun and the informal gender-neutral pronoun, 걔 (kyay) for a part of their gender-neutral corpus. However, for Hindi, we directly use the third person gender-neutral pronouns. This includes {\dn vh} (vah), {\dn v\?} (ve), {\dn vo} (vo) corresponding to formal impolite (familiar), formal polite (honorary) and informal (colloquial) respectively \cite{10.2307/30029211}.

As demonstrated by \citet{cho-etal-2019-measuring}, the performance of the MT system would be best evaluated with different sentence sets used as input. We apply the three categories of Hindi pronouns to make three sentence sets for each lexicon set (sentiment and occupations): (i) formal polite, (ii) formal impolite, and (iii) informal (colloquial use). 

\subsubsection{Evaluation}
We evaluate two systems, Google Translate and the Hi-En OpenNMT model, for seven lists that include: (a) informal, (b) formal, (c) impolite, (d) polite, (e) negative, (f) positive, and (g) occupation that are gender-neutral. We have attempted to find bias that exists in different types of contexts using these lists. The individual and cumulative scores help us assess contextual bias and overall bias in Hi-En translation respectively.

TGBI uses the number of translated sentences that contain she, he or they pronouns (and conventionally associated\footnote{The distinction between pronouns, gender and sex has been explain in section \ref{Ethics}} words such as girl, boy or person) to measure bias by associating that pronoun with $p_{he}$, $p_{she}$ and $p_{they}$\footnote{Changed convention to disassociate pronouns with gender and sex} for the scores of $P_{1}$ to $P_{7}$ corresponding to seven sets $S_{1}$ to $S_{7}$ such that:
\begin{equation}
    P_{i} = \sqrt{(p_{he}*p_{she}+p_{they})}
\end{equation}
 and finally, TGBI = avg($P_{i}$).

\section{Results and Discussion}
The BLEU score of the OpenNMT model we used was 24.53, and the RIBES score was 0.7357 across 2478 samples.
\subsection{WEAT}
We created multiple sets of categories for the attributes associated with `masculine' and `feminine,' including the subqueries as listed in the supplementary material. We used both the embeddings from the encoder and the decoder, that is to say, the source and the target embeddings, as the input to WEFE alongside the set of words defined in the target and attribute sets. Aside from this, we have also tested pre-trained word embeddings that were available with the gensim \cite{rehurek2011gensim} package on the same embeddings. The results of the measurement of bias using the WEFE framework are listed in Table \ref{WEFE Results}. 

For the English embeddings, there is a significant disparity in the WEAT measurement for the Math vs Arts and the Science vs Arts categories. This could be owing to the fact that there is little data in the corpus that the MT system was trained over, which is relevant to the attributes in these sets. Hence the bias is minimal compared to the pre-trained word2vec embeddings, which is learned over a dataset containing 100 billion words and is likely to learn more social bias compared to the embeddings learned in the training of the MT system. We notice a skew in some of the other results, which could be due to the MT model picking up on gender signals that have strong associations of the target set with the attribute set, implying a strong bias in the target set training data samples itself. However, all of these metrics and the pre-trained embeddings used are in positive agreement with each other regarding the inclination of the bias.

For the Hindi embeddings, while the values agree with each other for the first two metrics, there is a much more noticeable skew in the RND and ECT metrics. The pre-trained embeddings seem to exhibit much more bias, but the estimation of bias within the embedding learned by the MT may not be accurate due to the corresponding word vectors not containing as much information, consider the low frequency of terms in the initial corpus that the NMT was trained on. In addition to this, there were several words in the attribute sets in English that did not have an equivalent Hindi translation or produced multiple identical attribute words in Hindi. Consequently, we had to modify the Hindi attribute lists. 

While these metrics can be used to quantify gender bias, despite not necessarily being robust, as is illustrated in \citet{ethayarajh-etal-2019-understanding} which delves into the flaws of WEAT, they also treat gender in binary terms, which is also a consistent trend across research related to the field.

Our findings show a heavy tendency for Hi-En MT systems to produce gendered  outputs when the gender-neutral equivalent is expected. We see that many stereotypical biases are present in the source and target embeddings used in our MT system. Further work to debias such models is necessary, and the development of a more advanced NMT would be beneficial to produce more accurate translations to be studied for bias.

\subsection{TGBI}
The final TGBI score which is the average of different $P_{i}$ values, is between 0 and 1. A score of 0 corresponds to high bias (or gendered associations in translated text) and 1 corresponds to low bias \cite{cho-etal-2019-measuring}. 

The bias values tabulated in Table \ref{TGBI Results}, show that within both models, compared to the results on sentiment lexicons, occupations show a greater bias, with $p_{she}$ value being low. This points us directly to social biases projected on the lexicons ($S_{bias}$\footnote{In \citet{cho-etal-2019-measuring}, the authors describe two kinds of bias: $V_{bias}$ which is based on the volume of appearance in the corpora and $S_{bias}$ which is based on social bias that is projected in the lexicons. }). For politeness and impoliteness, we see that the former has the least bias and the latter most across all lists. While considering formal and informal lists, informal pronoun lists show higher bias. There are a couple of things to consider within these results: a) the polite pronoun {\dn v\?} (ve) is most often used in plural use in modern text ($V_{bias}$), thus leading to a lesser measured bias, b) consider that both polite and impolite are included in formal which could correspond to its comparatively lower index value compared to informal. 

Bias in MT outputs whether attributed to $S_{bias}$ or $V_{bias}$, is harmful in the long run. Therefore, in our understanding, the best recommendation is that TGBI = 1 with corresponding $p_{they}$, $p_{she}$, $p_{he}$ values 1, 0, 0 respectively.

\section{Bias Statement}
\subsection{Bias Statement}
In this paper, we examine gender bias in Hi-En MT comprehensively with different categories of occupations, sentiment words and other aspects. We consider bias as the stereotypical associations of words from these categories with gender or more specifically, gendered words. Based on the suggestions by \citet{blodgett2020language}, we have the two main categories of harms generated by bias: 1) representational, 2) allocational. The observed biased underrepresentation of certain groups in areas such as Career and Math, and that of another group in Family and Art, causes direct representational harm. Due to these representational harms in MT and other downstream applications, people who already belong to systematically marginalized groups are put further at risk of being negatively affected by stereotypes. Inevitably, gender bias causes errors in translation \cite{stanovsky-etal-2019-evaluating} which can contribute to allocational harms due to disparity in how useful the system proves to be for different people, as described in an example in \citet{savoldi2021gender}. The applications that MT systems are used to augment or directly develop increase the risks associated with these harms. 

There is still only a very small percent of the second most populated country in the world, India that speaks English, while English is the most used language on the internet. It is inevitable that a lot of content that might be consumed now or in the future might be translated. It becomes imperative to evaluate and mitigate the bias within MT systems concerning all Indic languages.

\subsection{Ethical Considerations and Suggestions}\label{Ethics}
There has been a powerful shift towards ethics within the NLP community in recent years and plenty of work in bias focusing on gender. However, we do not see in most of these works a critical understanding of what gender means. It has often been used interchangeably with the terms `female' and `male' that refer to sex or the external anatomy of a person. Most computational studies on gender see it strictly as a binary, and do not account for the difference between gender and sex. Scholars in gender theory define gender as a social construct or a learned association. Not accommodating for this definition in computational studies not only oversimplifies gender but also possibly furthers stereotypes \cite{brooke-2019-condescending}. It is also important to note here that pronouns in computational studies have been used to identify gender, and while he and she pronouns in English do have a gender association, pronouns are essentially a replacement for nouns. A person's pronouns, like their name, are a form of self-identity, especially for people whose gender identity falls outside of the gender binary \cite{article}. We believe research specifically working towards making language models fair and ethically sound should be employing language neutralization whenever possible and necessary and efforts to make existing or future methodologies more inclusive. This reduces further stereotyping \cite{Harris2017, Tavits16781}. Reinforcing gender binary or the association of pronouns with gender may be invalidating for people who identify themselves outside of the gender binary \cite{article}.

\section{Conclusion and Future Work}
In this work, we have attempted to gauge the degree of gender bias in a Hi-En MT system. We quantify gender bias (so far only for the gender binary) by using metrics that take data in the form of queries and employ slight modifications to TGBI to extend it to Hindi. We believe it could pave the way to the comprehensive evaluation of bias across other Indic and/or gendered languages. Through this work, we are looking forward to developing a method to debias such systems and developing a metric to measure gender bias without treating it as an immutable binary concept.

\section{Acknowledgements}
The authors of the paper are grateful for the contributions of Rianna Lobo for their reviews on the Bias Statement and Ethics Section. The efforts of the reviewers in reviewing the manuscript and their valuable inputs are appreciated. We would also like to thank the Research Society MIT for supporting the project.

\bibliography{anthology,custom}
\bibliographystyle{acl_natbib}
\end{document}